\documentclass[sigconf]{acmart}

\usepackage{booktabs}
\usepackage{adjustbox}
\usepackage{array}
\usepackage{multirow} 
\usepackage{placeins}
\usepackage{siunitx}
\usepackage{url}
\usepackage{stfloats}
\usepackage{caption}
\usepackage{cuted}
\usepackage{capt-of}
\newcolumntype{L}[1]{>{\raggedright\arraybackslash}p{#1}}

\setcopyright{rightsretained}\copyrightyear{2026}
\acmYear{2026}
\acmDOI{XXXXXXX.XXXXXXX}
\acmConference[CHASE]{Smart Connected Health}{Month DD--DD, 2026}{City, Country}
\acmISBN{978-1-4503-XXXX-X/26/XX}

\begin{document}

\title{Multimodal Routing for Interpretable, Robust, and Auditable Clinical Prediction}

\author{Nikkie Hooman}

\email{nikkieh@smu.edu}
\affiliation{%
  \institution{Southern Methodist University}
  \city{Dallas}
  \country{USA}
}

\author{Zhongjie Wu}
\authornotemark[1]
\email{zhongjiew@smu.edu}
\affiliation{%
  \institution{Southern Methodist University}
  \city{Dallas}
  \country{USA}
}

\author{Eric C. Larson}
\authornotemark[1]
\email{eclarson@smu.edu}
\affiliation{%
  \institution{Southern Methodist University}
  \city{Dallas}
  \country{USA}
}

\author{Mehak Gupta}
\authornotemark[1]
\email{mehakg@smu.edu}
\affiliation{%
  \institution{Southern Methodist University}
  \city{Dallas}
  \country{USA}
}
\begin{abstract}
EHR data are inherently multimodal, and leveraging multiple modalities can improve predictive performance. However, most existing approaches rely on deep fusion, which obscures how individual modalities contribute to predictions and limits the interpretability of multimodal reasoning. We propose an explicit multimodal routing framework for clinical prediction that enables interpretable, robust, and auditable reasoning across three EHR modalities: structured longitudinal variables ($L$), clinical notes ($N$), and chest X-rays ($I$). Our model constructs discrete unimodal, directional bimodal, and trimodal routes to capture both individual modality signals and asymmetric cross-modal interactions. To audit multimodal reasoning and assess robustness, we introduce inference-time route masking, which simulates missing modalities and reweights the remaining routes without retraining. We analyze changes in performance and routing weights under these scenarios to understand model decision-making. We evaluate our framework on multi-label phenotype prediction ($K=25$) and binary ICU mortality prediction using trimodal patient stays from MIMIC-IV, revealing systematic differences in modality reliance across clinical condition groups. Overall, our framework offers a transparent, auditable, and practical approach to multimodal clinical prediction, providing interpretability, robustness, and insights into how different data sources drive model decisions.

\end{abstract}




\maketitle

\section{Introduction}

Clinicians in Intensive Care Units (ICUs) routinely make high-stakes decisions by integrating heterogeneous information, including longitudinal vital signs and laboratory measurements, clinical notes, and medical imaging. These diverse data sources are captured within Electronic Health Records (EHRs), creating a strong motivation for multimodal AI (MAI) systems that jointly model structured variables, free text, and images. Compared to unimodal approaches, multimodal models can leverage complementary clinical evidence, improve predictive performance, and better tolerate noisy or incomplete inputs \cite{lee2023learning, wang2025integrating}. By supporting structured multimodal reasoning that reflects how clinicians combine evidence across sources, multimodal systems offer a strong foundation for accurate and clinically meaningful decision support.

Recent studies have demonstrated consistent performance gains from multimodal models (MAI) across a range of clinical prediction tasks \cite{huang2020fusion, patil2025multimodal, wang2025robust, sun2024outcome}. However, most MAI systems rely on deep fusion strategies that combine modalities into a single latent representation, rendering model behavior difficult to interpret. This opacity is particularly problematic in clinical settings, where trust depends not only on predictive accuracy but also on understanding which sources of evidence drive a given prediction. Because the relevance of each modality varies across clinical outcomes, effective multimodal systems must move beyond a single fixed fusion strategy and instead adaptively combine modalities in a transparent and task-aware manner. Without such transparency, even high-performing MAI models may struggle to gain clinical acceptance.

To address these challenges, we propose a multimodal routing framework that explicitly separates and evaluates the contributions of structured laboratory data ($L$), clinical notes ($N$), and medical images ($I$), as well as their interactions. The framework defines ten explicit routes corresponding to unimodal, directional bimodal, and trimodal information flow. By treating routes such as $A \leftarrow B$ and $B \leftarrow A$ as distinct, the model captures asymmetric cross-modal effects that more closely align with clinical reasoning processes. Learnable routing weights quantify the contribution of each route to the final prediction, enabling both local and global interpretability.

We evaluate our approach using trimodal ICU stays from the MIMIC-IV dataset on two clinically relevant tasks: binary ICU mortality prediction and multi-label phenotype prediction across 25 conditions. Beyond predictive performance, we analyze routing activations and coefficients to characterize how evidence is combined across tasks. We further assess model robustness through inference-time route masking, simulating missing-modality scenarios, and examining how routing weights redistribute under incomplete information. This analysis provides an auditable view of model behavior and reliability under realistic clinical conditions. Our contributions include:

\begin{itemize}
\item  We introduce a routing-based multimodal framework that explicitly models unimodal, directional bimodal, and trimodal interactions, capturing cross-modal clinical reasoning.

\item We implement routing activations and coefficients to quantify the contribution of each interaction, providing local and global interpretability across 25 phenotypes.

\item  We analyze changes in routing weights and performance under missing-modality scenarios to evaluate robustness and audit the model’s multimodal reasoning.

\end{itemize}

\section{Related Work}
\subsection{Multimodal Fusion in EHR}

A substantial body of work has explored integrating structured Electronic Health Record (EHR) data with unstructured modalities such as clinical notes and medical images. For example, MINGLE models structured EHR variables as a hypergraph and injects information from other modalities to learn a shared patient representation \cite{cui2024multimodal}. Other approaches employ multimodal transformers to jointly encode clinical time series and notes, fusing information via self-attention across the ICU timeline for mortality prediction \cite{lyu2023multimodal}, or utilize LSTM-based joint fusion modules to aggregate sequence and imaging data, as seen in MedFuse \cite{hayat2022medfuse}. While these methods demonstrate strong performance gains, they rely on deep fusion mechanisms that collapse multiple modalities into a single latent space, making it difficult to determine whether a prediction is driven by a specific modality, cross-modal interactions, or spurious correlations. Late fusion approaches partially address this interpretability bottleneck by combining modality-specific models at the decision level. For instance, \citet{ruan2025towards} fuse structured data and clinical notes via late fusion, and RadFusion integrates CT imaging with EHR data for pulmonary embolism detection \cite{zhou2021radfusion}. Although these methods permit explicit analysis of unimodal contributions, they combine modalities at the decision level and do not capture how modalities interact to jointly influence predictions.

A more recent line of work explicitly models multimodal interactions rather than treating fusion as a purely additive process \cite{wortwein2022beyond, tsai2020multimodal}. TriMF \cite{wang2025missing} constructs three bimodal pathways but sums them into a single multimodal embedding, collapsing the pathway structure before prediction. Our framework instead preserves each unimodal, directional bimodal, and trimodal route as a separately parameterized pathway and supports inference-time weighting, enabling post-hoc auditing of multimodal reasoning without retraining.

Mixture-of-Experts (MoE) architectures form another prominent direction for multimodal fusion under missing-modality conditions. MoE-Health pretrains experts on specific modality combinations and applies gating to the top-$k$ routes \cite{wang2025moe}. FuseMoE introduces Laplace-gated sparse MoE fusion over modality embeddings, designed for flexible handling of missing modalities and irregular temporal sampling \cite{han2024fusemoe}. Flex-MoE replaces transformer feed-forward layers with sparse MoE routing over modality embeddings supplemented by a learnable missing-modality bank \cite{yun2024flex}. A related approach, FAME, applies fairness-aware scalar weighting over unimodal embeddings \cite{hooman2025equitable}. These methods fuse modalities by concatenation, weighted sum, or stacked self-attention before producing predictions, and do not expose directional cross-modal pathways. Consequently, the contribution of an interaction such as ``notes conditioned on imaging'' versus ``imaging conditioned on notes'' cannot be individually inspected or ablated, which is the mechanism underlying our route-level auditing framework.

\subsection{Interpretability and Auditable Multimodal Reasoning}

Explicit modeling of multimodal interactions naturally enables interpretability and auditability—key requirements for clinical decision support. In contrast to deep fusion approaches, where modality contributions are entangled, routing-based interaction modeling exposes distinct evidence pathways that align with clinical reasoning. Learnable routing coefficients quantify the contribution of each unimodal and cross-modal route, supporting both local interpretation at the patient level and global analysis across phenotypes.

Prior multimodal approaches, such as Mixture-of-Experts methods discussed above, have addressed modality missingness to preserve predictive performance \cite{yao2024drfuse, al2025multimodal}. Although effective, these methods provide limited insight into how the availability or removal of specific cross-modal interactions affects model decision-making.

In our framework, missing-modality analysis serves as an auditing mechanism rather than solely a robustness strategy. By masking routes associated with missing modalities at inference and renormalizing over the remaining routes, we directly observe how the model redistributes reliance across evidence pathways. Changes in routing weights and performance reveal whether predictions are supported by complementary modalities or dominated by narrow, potentially unreliable signals. This allows clinicians and researchers to assess the reliability of automated predictions based on the structure of multimodal evidence supporting them.

Overall, by linking multimodal interaction modeling with explicit routing and inference-time auditing, our approach provides a transparent and clinically grounded framework for interpretable and auditable multimodal prediction in EHR data.

\section{Methods}

\subsection{Problem Formulation}
We design a multimodal AI (MAI) framework that jointly models three clinical modalities—structured longitudinal data ($L$), clinical notes ($N$), and medical images ($I$)—and is trained on multiple supervised clinical prediction tasks. To handle these heterogeneous modalities in a unified manner, we represent each modality as a sequence of tokens or feature units. For modality $M \in \{L, N, I\}$, the input tokens are denoted as $X_M = \{x_1, x_2, \ldots, x_{T_M}\}$, where $T_M$ is the number of tokens for that modality. For structured data, tokens correspond to time-ordered clinical events (e.g., vitals or lab measurements), while for clinical notes, tokens correspond to words or subwords in the text. For medical images, tokens are derived from a convolutional backbone (e.g., ResNet) and correspond to spatial feature regions of the image.

Each modality is processed by a modality-specific encoder ($enc_M$) to capture the unique structure of each data type, producing hidden representations
\begin{equation}\label{eq:1}
H_M=enc_M(X_M)
\end{equation}
where $H_M \in \mathbb{R}^{B_{\text{sz}} \times T_M \times d_M}$, $B_{\text{sz}}$ is the batch size, and $d_M$ is the encoder-specific hidden dimension. Embeddings are pooled across the dimension $T_M$ using a pooling operator 
(e.g., CLS token selection, mean pooling, or max pooling) to obtain a single global vector summarizing the entire modality, $z_M\in\mathbb{R}^{B_{\text{sz}}\times d_M}$. Since the number of tokens ($T_M$) may vary across patients, we define a binary mask,
$M_M \in \{0,1\}^{B_{\text{sz}} \times T_M}$
to indicate valid tokens (non-padding) that contribute towards pooling. 

Different encoders produce latent representations with modality-specific dimensionalities. To align these representations, we apply a linear projection to each modality’s pooled output:
\begin{equation}
\tilde{z}_M = z_M W_M + b_M, \quad \tilde{z}_M \in \mathbb{R}^{B_{\text{sz}} \times d},
\end{equation}
where $z_M \in \mathbb{R}^{B_{\text{sz}} \times d_M}$ is the pooled encoder output for modality $M$, 
$W_M \in \mathbb{R}^{d_M \times d}$ is a learnable projection matrix, and 
$b_M \in \mathbb{R}^{d}$ is a bias vector. The projected representations $\tilde{z}_M$ are then used in subsequent fusion and cross-modal interaction modules.

\begin{figure*}[t]
  \centering
  \includegraphics[width=\textwidth]{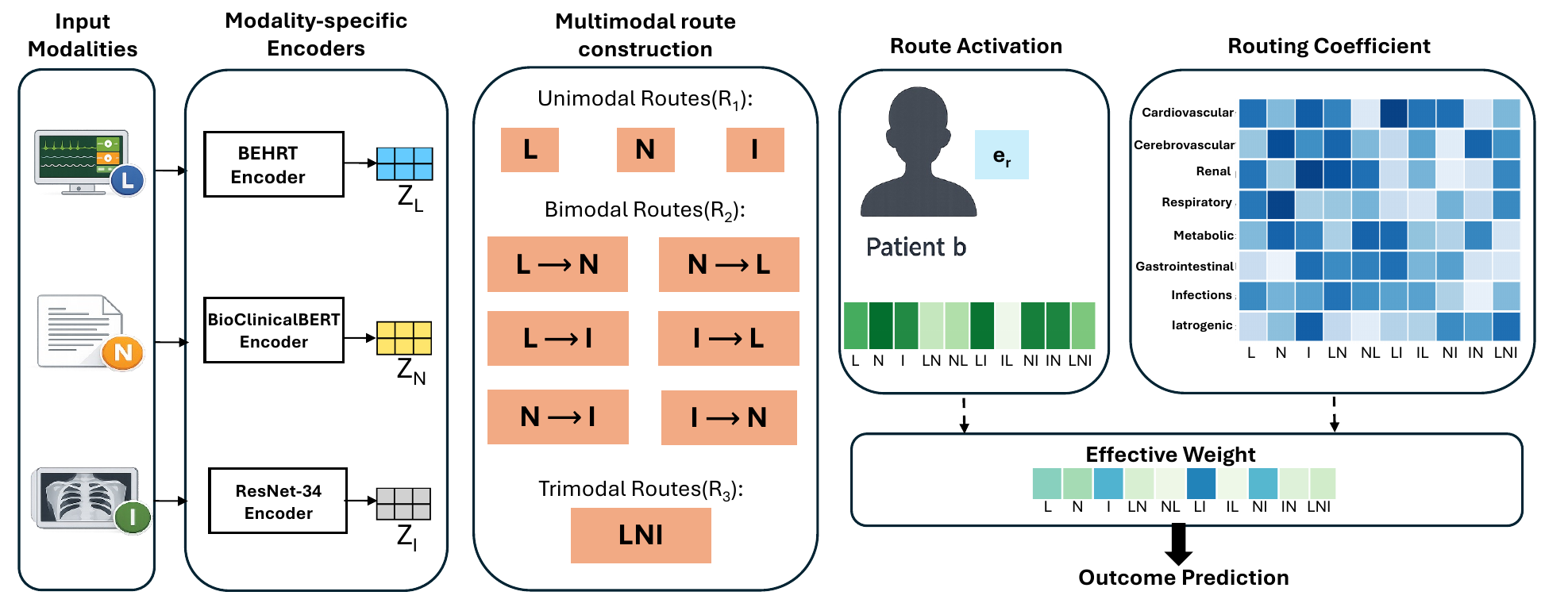}
  \caption{\textbf{Architecture of the proposed multimodal routing framework.}}
  \label{fig:architecture}
\end{figure*}

In the following sections, we detail each component of the proposed framework, as illustrated in Figure~\ref{fig:architecture}.

\subsection{Modality-Specific Encoders}

\paragraph{Structured encoder.}
The structured modality consists of longitudinal sequences of vitals, medications, and other measurements aggregated in hourly bins. Each clinical event is embedded into a token representation, and positional embeddings are added to maintain longitudinal order. We pass the sequence through a BEHRT Transformer \cite{li2020behrt}. Let the token-level embeddings for a patient be
\[
H_L = [h_{L,1}, \ldots, h_{L,T_L}] \in \mathbb{R}^{B_{\text{sz}} \times T_L \times d_L},
\]
where $T_L$ is the number of hourly bins and $d_L$ is the token embedding dimension.  
We obtain the patient-level embedding by taking the \texttt{[CLS]} token output from the last layer:
\[
z_L \in \mathbb{R}^{B_{\text{sz}} \times d_L}.
\]

\paragraph{Clinical notes encoder.}
Clinical notes within the observation window are concatenated in chronological order and split into overlapping chunks of length $L=512$ tokens. Each chunk is passed through BioClinicalBERT \cite{alsentzer2019publicly} to obtain chunk-level embeddings
\[
h_{N,T_s} \in \mathbb{R}^{B_{\text{sz}} \times d_N}
\]
where $T_S$ is the number of chunks. Stacking the chunk embeddings gives the token-level representation
\[
H_N = [h_{N,1}, \ldots, h_{N,T_S}] \in \mathbb{R}^{B_{\text{sz}} \times T_S \times d_N}.
\]
We then compute a patient-level embedding by mean pooling over chunks:
\[
z_N = \frac{1}{S} \sum_{s=1}^{S} h_{N,s} \in \mathbb{R}^{B_{\text{sz}} \times d_N}.
\]

\paragraph{Image encoder.}
Chest X-ray images are resized to $224 \times 224$ and passed through a ResNet-34 \cite{he2016deep, hayat2022medfuse}. From the final convolutional block, we extract spatial feature maps and flatten them into image tokens
\[
H_I = [h_{I,1}, \ldots, h_{I,T_I}] \in \mathbb{R}^{B_{\text{sz}} \times T_I \times d_I},
\]
where $T_I$ is the number of spatial locations. To obtain a patient-level embedding, we apply global average pooling followed by a linear projection:
\[
z_I \in \mathbb{R}^{B_{\text{sz}} \times d_I}.
\]

\subsection{Multimodal Route Construction}
\label{sec:routes}
We build explicit routes that represent unimodal information and cross-modal interaction features, so we can track how each modality and each interaction contributes to the model decision. Our approach does not assume multimodal interactions to be symmetric because conditioning modality \(A\) on modality \(B\) can capture different evidence than conditioning modality \(B\) on modality \(A\). Therefore, we construct these interactions (routes) with directional cross-attention, where we treat each direction as a separate route, and the model can learn to weigh each route differently when forming the final representation. %\cite{tsai2019multimodal}.
\paragraph{Unimodal routes (\(\mathcal{R}_1\)).}
Each modality encoder returns a patient-level embedding that summarizes a single modality information for that patient, and we use these embeddings as unimodal route vectors. Each unimodal embedding projected into a \(d\)-dimensional representation forms the unimodal set $\mathcal{R}_1=\{L, N, I\}$.

\paragraph{Bimodal routes (\(\mathcal{R}_2\)).}

We define a bimodal block that constructs a directed interaction route for every ordered pair of modalities. For a bimodal route $A \leftarrow B$, modality $A$ is updated using contextual information from modality $B$ via cross-attention.

Each modality is encoded using a modality-specific encoder to produce token-level embeddings, projected into a shared latent space of dimension $d$,
\[
H_A \in \mathbb{R}^{B_{\text{sz}} \times T_A \times d}, H_B \in \mathbb{R}^{B_{\text{sz}} \times T_B \times d}
\]

To model directed interactions $A \leftarrow B$, we use token-level embeddings for A as queries ($Q = H_A W_Q$) and for B as keys ($K = H_B W_K$) and values ($V = H_B W_V$), where $W_Q, W_K, W_V \in \mathbb{R}^{d \times d_k}$ are learned projection matrices. Cross-attention is then computed as
\[
\widehat{H}_{A\leftarrow B} = \mathrm{softmax}\!\Big(\frac{QK^\top}{\sqrt{d_k}}\Big) V \in \mathbb{R}^{B_{\text{sz}} \times T_A \times d}.
\]

We apply a residual connection and feed-forward network to update the tokens of modality A:
\[
\widetilde{H}_{A\leftarrow B} = \mathrm{FFN}(H_A + \widehat{H}_{A\leftarrow B}) \in \mathbb{R}^{B_{\text{sz}} \times T_A \times d}.
\]

Finally, we obtain the fixed-dimensional route embedding by mean pooling over the updated tokens of A:
\begin{equation}
e_{A\leftarrow B} = \frac{1}{T_A} \sum_{t=1}^{T_A} \widetilde{H}_{A\leftarrow B}[:, t, :] \in \mathbb{R}^{B_{\text{sz}} \times d}.
\end{equation}

This formulation produces a route-level embedding $e_{A\leftarrow B}$ that summarizes how modality $B$ contextualizes modality $A$.

This block is reused for all ordered modality pairs \(A, B \in \{L, N, I\}\), enabling the construction of six directional bimodal routes, $\mathcal{R}_2=\{L\!\leftarrow\!N,\ N\!\leftarrow\!L,\ L\!\leftarrow\!I,\ I\!\leftarrow\!L,\ N\!\leftarrow\!I,\ I\!\leftarrow\!N\}$.

\paragraph{Trimodal route (\(\mathcal{R}_3\)).}
Trimodal reasoning is modeled as a structured composition of pairwise cross-modal interactions. 
From the bimodal routing block, we obtain six directional embeddings:

\[
\{ e_{L\leftarrow N}, e_{N\leftarrow L}, e_{L\leftarrow I}, e_{I\leftarrow L},
   e_{N\leftarrow I}, e_{I\leftarrow N} \} \in \mathbb{R}^{B_{\text{sz}} \times d}.
\]
We first form three modality-pair representations by combining each directional pair:
\begin{align}
e_{LN} &= \mathrm{Proj}_{LN}\big([e_{L\leftarrow N}; e_{N\leftarrow L}]\big), \\
e_{LI} &= \mathrm{Proj}_{LI}\big([e_{L\leftarrow I}; e_{I\leftarrow L}]\big), \\
e_{NI} &= \mathrm{Proj}_{NI}\big([e_{N\leftarrow I}; e_{I\leftarrow N}]\big),
\end{align}
where $[\,;\,]$ denotes concatenation and $\mathrm{Proj}_{*}(x)$ is a learnable linear projection mapping from $\mathbb{R}^{2d}$ to $\mathbb{R}^{d}$. The trimodal route is then constructed by combining the three modality-pair embeddings:
\begin{equation}
e_{LNI} = \mathrm{Proj}_{T}\big([e_{LN}; e_{LI}; e_{NI}]\big),
\end{equation}
where $e_{LNI} \in \mathbb{R}^{B_{\text{sz}}\times d}$, and $\mathrm{Proj}_{T}$ projects $\mathbb{R}^{3d}$ to $\mathbb{R}^{d}$.

This hierarchical formulation grounds trimodal reasoning in structured combinations of directional cross-modal evidence, while preserving interpretability and enabling analysis of how multimodal interactions contribute to prediction.

\paragraph{Final route set.}
Putting everything together, we get 3 unimodal routes \(\mathcal{R}_1\), 6 directional bimodal routes \(\mathcal{R}_2\), and 1 trimodal route \(\mathcal{R}_3\) giving a total of 10 routes
\begin{equation}
\label{eq:Rset}
\mathcal{R}=\mathcal{R}_1\cup\mathcal{R}_2\cup\mathcal{R}_3,
\qquad |\mathcal{R}|=10.
\end{equation}

\subsection{Routing Activations and Routing Coefficients}

Each route embedding $e_r \in \mathbb{R}^{B_{\text{sz}}\times d}$ captures information from a specific unimodal, bimodal, or trimodal interaction.  
However, not every route is equally relevant for every patient or outcome.  
We therefore introduce two complementary quantities: 
(i) \emph{route activations}, which measure how strongly a route is expressed for a given patient, independent of any specific prediction target,
and  
(ii) \emph{routing coefficients}, quantify how much each route contributes to each prediction target.
Together, they ensure that only routes both strongly expressed and relevant for the target influence the decision, enforcing structured aggregation of multimodal information.

\paragraph{Route Activations.}
Each route embedding $e_r$ is projected into a route vector of size $a+1$.  
The first $a$ dimensions form the route vector $\mathbf{\tilde e}_r \in \mathbb{R}^{B_{\text{sz}} \times a}$, carries the predictive information,  
and the last dimension is the sigmoid activation, which acts as the route activation $\alpha_{p,r} \in (0,1)$, indicating how strongly route $r$ is expressed for patient $p$. These activations are patient-specific indicators allowing the model to downweight weak routes. 

\paragraph{Routing Coefficients.}
The routing layer produces patient- and label-specific routing coefficients $R_{p,r,c}$, normalized over routes for each patient $p$ and label $c$ such that
 $\sum_{r \in \mathcal{R}} R_{p,r,c} = 1$. A high $R_{p,r,c}$ indicates strong agreement between route $r$ and label $c$ for patient $p$, whereas low values indicate less contribution.  
These coefficients act as dynamic weights that modulate the influence of each route on a specific outcome, while also enabling direct inspection of modality reliance and interaction patterns at both local (per-patient) and global (across dataset) levels.

\paragraph{Outcome Prediction.} 
The weighted aggregation of route embeddings gives the decision representation for patient $p$ for label $c$:
\begin{equation}
\mathbf{d}_{p,c} = \sum_{r \in \mathcal{R}} R_{p,r,c} \, \alpha_{p,r} \, \mathbf{\tilde e}_r.
\end{equation}
Here, $\mathbf{\tilde e}_r$ is the route content vector, $\alpha_{p,r}$ is the route activation, and $R_{p,r,c}$ is the routing coefficient.  
Final predictions are obtained by applying a $Softmax$ for binary tasks or a $Sigmoid$ for multi-label tasks.  
This design enforces explicit, structured, and interpretable contributions from each route while producing task-specific predictions.

\subsection{Robustness and Auditable Reasoning Under Missing Modalities}

Robustness to missing modalities is critical for multimodal clinical models, especially in ICU settings where data availability is often variable. Our framework explicitly models unimodal, directional bimodal, and trimodal routes, allowing predictions to rely on available evidence rather than failing under partial observability.

To evaluate robustness, we mask modalities at inference time. All routes involving the masked modality are disabled, and the routing coefficients of the remaining active routes are reweighted. The model is not retrained, so this evaluation reflects performance under post-deployment missing-modality scenarios.

This approach also provides a natural audit of multimodal reasoning. By examining how routing weights and performance change when modalities are missing, we can assess how the model redistributes reliance across available evidence pathways. Stable performance supported by complementary routes indicates robust and clinically coherent reasoning, whereas large drops in performance or shifts toward narrow unimodal pathways reveal fragile or potentially unreliable decision logic.

\subsection{Interpretability}
By explicitly separating patient-specific route activations from task-specific routing coefficients, our framework supports interpretability of multimodal clinical predictions. By combining activations $\alpha_{p,r}$ and routing coefficients $R_{p,r,c}$, we define an effective contribution weight for each route and label it as 
$W_{p,r,c} = \alpha_{p,r} \cdot R_{p,r,c}$, 
capturing both route expression and task-specific importance. 
We use these variables to provide structured interpretability across individual patients (local), the full dataset (global), and disease-specific.

\paragraph{Local interpretation.}  
For an individual patient, we inspect route activations, routing coefficients, and their effective contributions to identify which unimodal or cross-modal interactions dominate the prediction for different labels. This gives patient-level insight into modality reliance and the reasoning behind the model’s decision.

\paragraph{Global interpretation.}  
At the dataset level, we average route activations, routing coefficients, and effective weights across all patients to assess overall modality usage for each disease label. This will help quantify which routes drive predictions across the dataset.  

\paragraph{Group-level interpretation.}  
To analyze patterns across clinically related outcomes, we group labels into disease categories (e.g., all heart-related conditions, all lung-related conditions). Effective contributions are averaged across patients within each disease group to reveal which routes are most prominent for specific types of outcomes. This provides a more nuanced, group-level view of the model’s structured multimodal reasoning specific to each disease group.

\section{Experiment Design}
\subsection{Cohort and Prediction Tasks}
We evaluate our proposed method on the MIMIC-IV EHR dataset \citep{johnson2023mimic}, which contains ICU stays with three modalities: structured longitudinal data ($L$), clinical notes ($N$) from the MIMIC-IV-Note module \citep{PhysioNet-mimic-iv-note-2.2}, and chest X-ray images ($I$) from MIMIC-CXR \citep{PhysioNet-mimic-cxr-jpg-2.1.0}.

We test on two prediction tasks - ICU mortality and phenotype prediction. ICU mortality prediction is a binary task, where we use paired EHR data for all modalities from the first 48 hours of the ICU stay ($N = 6,677$). Phenotype prediction is a multi-label task ($K=25$), where we identify 25 clinically meaningful phenotypes derived from ICD-9 and ICD-10 diagnosis codes assigned during the ICU stay, following established phenotype groupings used in prior MIMIC-based benchmark studies \citep{harutyunyan2019multitask}. We use full-stay paired EHR data ($N = 11{,}230$). We exclude discharge summaries because they directly contain phenotype information, leading to data leakage.

The dataset is split into 80\% training and 20\% test sets, with 5\% of the training data reserved for validation. We report all results on the test set using the best model selected via early stopping on the validation set. Additional details on cohort construction and preprocessing are provided in Appendix~\ref{app:cohort_details}.

\subsection{Implementation Details}
All models are implemented in PyTorch. Structured EHR time series are encoded using a BEHRT-style Transformer, clinical notes are encoded using BioClinicalBERT, and chest X-ray images are encoded using a ResNet-34 backbone.

All modality encoders project to a shared embedding space of dimension $d=256$. Route activation representations use a dimension of $a=128$, and the final decision representations used for classification used a dimension $256$. Models are trained using the Adam optimizer with a learning rate of $1\times10^{-4}$ and a batch size of 16. Training proceeds for up to 50 epochs with early stopping based on validation loss.

We report performance using AUROC and F1. For the multi-label phenotype prediction task, we report macro-averaged F1 across the 25 phenotypes. All weight analyses are conducted using normalized weights across different multimodal routes.
Our open-source implementation is publicly available. \url{https://anonymous.4open.science/r/MultimodalRouting-85C1/README.md}

\subsection{Types of Experiments}

\subsubsection{Baseline Comparison.} We compare our routing-based model with multiple multimodal baseline techniques and also with more sophisticated multimodal models: 
\begin{itemize}
\item Joint fusion (average modality embeddings)
\item Late fusion \cite{huang2020fusion}
    \item TriMF \cite{wang2025missing}, which implements bimodal pathways and combines them to create a final multimodal embedding,
    \item MoE \cite{wang2025moe}, which routes inputs to modality-subset experts using gating-based selection.
\end{itemize}

These baselines cover the dominant fusion paradigms relevant to our setting: simple aggregation (joint and late fusion, also adopted by MedFuse \cite{hayat2022medfuse}, RadFusion \cite{zhou2021radfusion}, and \citet{ruan2025towards}), multi-pathway fusion (TriMF), and gated expert fusion (MoE-Health). Graph- and hypergraph-based fusion methods, such as MINGLE \cite{cui2024multimodal}, target a different input representation and are not directly comparable in our token-based encoder setting. We do not benchmark against methods whose central mechanisms target cohorts with varying modality availability, such as FuseMoE \cite{han2024fusemoe}, Flex-MoE \cite{yun2024flex}, and DrFuse \cite{yao2024drfuse}, as they do not apply to our fully-trimodal training data. All baselines use the same modality-specific encoders, training data, and optimization settings, and differ only in the fusion mechanism.

\subsubsection{Ablation Analysis.} We further conduct an ablation study to assess the contribution of key model components. Specifically, we evaluate our model under three settings: (i) removing routing activations ($\alpha$), (ii) removing routing coefficients ($r$), and (iii) removing cross-attention. In this setting, instead of modeling six directional bimodal interactions, we simply concatenate unimodal embeddings to form three undirected bimodal representations that correspond to the symmetric setting. 

% \begin{itemize}
%     \item Removing routing activations ($\alpha$)
%     \item Removing routing coefficients ($r$)
%     \item Removing cross-attention. In this setting, instead of modeling six directional bimodal interactions, we simply concatenate unimodal embeddings to form three undirected bimodal representations that correspond to the symmetric setting. 
% \end{itemize}
%(1) removing routing activations ($\alpha$), (2) removing routing coefficients ($r$), and (3) removing cross-attention. In the third setting, instead of modeling six directional bimodal interactions via cross-attention, we simply concatenate unimodal embeddings to form three undirected bimodal representations.

\subsubsection{Missing Modality Evaluation.}
We audit multimodal reasoning and robustness under missing-modality settings to assess model reliability in realistic deployment scenarios. For unimodal evaluation, each modality is assessed independently; for bimodal settings, one modality is masked at a time, and all routes involving the masked modality are removed at inference. Structured data are always retained, as they are consistently available in ICU environments and provide core clinical signals (vitals and laboratory measurements), making them a natural anchor modality. Masking structured data would therefore be clinically unrealistic and less informative. We analyze changes in predictive performance alongside shifts in route activations and routing coefficients to characterize how the model’s reasoning pathways adapt under missing modalities.

\subsubsection{Interpretability Analysis.}
Finally, we conduct interpretability analyses of routing activations and coefficients under both full- and missing-modality settings to examine how routing behavior and reasoning pathways influence performance and reliability. All analyses are performed using normalized routing weights across multimodal routes to ensure comparability.

\begin{table}[t]
\centering
\caption{Baseline comparison on the test set. We compare our model with the existing state-of-the-art multimodal fusion techniques using AUROC and F1. All Phenotype results are averages across 25 phenotypes. Best results are in bold.}
\label{tab:baseline_compare}
 \small
\begin{tabular}{l|cc|cc}
\toprule
& \multicolumn{2}{c|}{Mortality} & \multicolumn{2}{c}{Phenotypes} \\
Model & AUROC & F1 & AUROC & F1 \\
\midrule
%Concat fusion & 0.7875 & 0.4248 & 0.7583 & 0.4895 \\

Joint fusion & $0.70_{\pm0.02}$ & $0.34_{\pm0.02}$ & $0.69_{\pm0.03}$ & $0.37_{\pm0.02}$ \\
Late Fusion & $0.72_{\pm0.01}$ & $0.35_{\pm0.02}$ & $0.71_{\pm0.02}$ & $0.40_{\pm0.01}$ \\
TriMF & $0.75_{\pm0.01}$ & $0.41_{\pm0.01}$ & $0.72_{\pm0.01}$ & $0.41_{\pm0.02}$ \\
MoE-Health & $0.77_{\pm0.01}$ & $0.43_{\pm0.01}$ & $0.75_{\pm0.02}$ & $0.47_{\pm0.02}$ \\\hline

Multimodal Routing (ours) &  $\mathbf{0.80_{\pm0.02}}$ & $\mathbf{0.45_{\pm0.01}}$  & $\mathbf{0.78_{\pm0.01}}$ & $\mathbf{0.50_{\pm0.01}}$ \\
\bottomrule
\end{tabular}
\end{table}

\begin{table}[t]
\centering
\caption{Ablation analysis on the test set. We compare our multimodal routing model with different ablations by removing one component at a time. We evaluate performance using AUROC and F1. All Phenotype results are averages across 25 phenotypes. Best results are shown in bold.}
\label{tab:ablation}
 \small
\begin{tabular}{l|cc|cc}
\toprule
& \multicolumn{2}{c|}{Mortality} & \multicolumn{2}{c}{Phenotypes} \\
Ablations & AUROC & F1 & AUROC & F1 \\
\midrule
%Concat fusion & 0.7875 & 0.4248 & 0.7583 & 0.4895 \\

w/o cross-attention & $0.75_{\pm0.02}$ & $0.40_{\pm0.02}$ & $0.73_{\pm0.02}$ & $0.46_{\pm0.01}$ \\
w/o Route Activations & $0.71_{\pm0.02}$ & $0.36_{\pm0.03}$ & $0.74_{\pm0.02}$ & $0.45_{\pm0.02}$ \\
w/o Routing Coeff. & $0.72_{\pm0.01}$ & $0.37_{\pm0.01}$ & $0.75_{\pm0.02}$ & $0.47_{\pm0.01}$ \\
\midrule
Full Multimodal Routing &  $\mathbf{0.80_{\pm0.02}}$ & $\mathbf{0.45_{\pm0.01}}$  & $\mathbf{0.78_{\pm0.01}}$ & $\mathbf{0.50_{\pm0.01}}$ \\
\bottomrule
\end{tabular}
\end{table}

\section{Results}
\label{sec:results}
\subsection{Predictive Performance}
\label{sec:results_metrics}

\subsubsection{Baselines Comparison.} 
Table~\ref{tab:baseline_compare} reports performance on ICU mortality and multi-label phenotype prediction, comparing our approach with established multimodal fusion baselines and targeted ablations of our routing framework. For phenotype prediction, results are averaged across 25 phenotypes.

Among baseline methods, more advanced multimodal methods such as MoE and TriMF outperform simple joint and late fusion baselines, confirming the benefit of modeling multimodal structure. However, our routing-based method achieves the best performance, improving ICU mortality AUROC and F1 by 3.9\% and 4.7\%, and phenotype prediction AUROC and F1 by 4.0\% and 6.4\% over the next-best model, highlighting the benefits of explicitly modeling unimodal and cross-modal interactions.

\subsubsection{Ablation Analysis.} 
The ablation results in Table \ref{tab:ablation} further clarify the role of each component in enabling structured multimodal reasoning. Removing routing activations or coefficients leads to large performance degradation (17-20\% in F1), confirming that these components form the core of our framework by explicitly and systematically weighting multimodal evidence.  Removing cross-attention also degrades performance (11\% on F1), demonstrating that directional modeling of bimodal interactions strengthens the routing framework by distinguishing how information flows between modalities, rather than assuming symmetric effects. These results show that directional cross-attention, patient-specific route weighting via routing activations, and label-specific weighting through routing coefficients play complementary roles in enabling robust, selective, and interpretable multimodal reasoning.

\subsubsection{Per-Phenotype Comparison.} Table~\ref{tab:per_pheno_grouped_1col_auroc_f1_prev} summarizes per phenotype performance, along with prevalence and group-level averages. Overall, predictive performance varies substantially across phenotypes and disease groups, reflecting differences in prevalence, signal availability, and clinical heterogeneity.

Across groups, cardiovascular, renal, metabolic, and infectious phenotypes achieve consistently strong performance (mean $F_1$ between 0.52 and 0.56), indicating that these conditions are well supported by multimodal EHR signals. In particular, highly prevalent ($prev=$0.30 to 0.43) phenotypes such as essential hypertension, fluid and electrolyte disorders, lipid metabolism disorders, congestive heart failure, and renal failure demonstrate strong predictive performance ($F_1=$ 0.61 to 0.67), consistent with their reliance on structured longitudinal measurements and well-documented clinical patterns.

Respiratory and gastrointestinal phenotypes exhibit lower average performance (mean $F_1 =$ 0.38 to 0.40), driven largely by low-prevalence and heterogeneous conditions such as upper and lower respiratory disease ($prev=$ 0.06 to 0.11) and gastrointestinal hemorrhage ($prev=$ 0.06). These results highlight the challenge of learning reliable predictors under class imbalance and variable clinical presentation, even with multimodal input.

Notably, acute cerebrovascular disease achieves relatively high performance ($F_1 = 0.61$) despite low prevalence (0.07). This deviation from the general prevalence–performance trend motivates deeper interpretability and auditing analyses to assess whether high accuracy in low-prevalence settings reflects robust multimodal evidence or reliance on spurious correlations.

\begin{table}[t]
\centering
\caption{We report the per-phenotype performance (F1) on the test dataset using our proposed model. We also group each phenotype into disease categories and report the mean performance across each group.}
\label{tab:per_pheno_grouped_1col_auroc_f1_prev}
\small

\setlength{\tabcolsep}{3pt}
\renewcommand{\arraystretch}{0.85}

\resizebox{\columnwidth}{!}{%
\begin{tabular}{l cccc}
\toprule
Phenotype  & F1 (Full) & F1 (Missing I) & F1 (Missing N) & Prev \\
\midrule

\multicolumn{5}{l}{\textbf{Cardiovascular}}\\
Acute myocardial infarction  & 0.28 & 0.23 & 0.23 & 0.09 \\
Cardiac dysrhythmias  & 0.57 & 0.56 & 0.55 & 0.37 \\
Conduction disorders  & 0.57 & 0.51 & 0.29 & 0.10 \\
Congestive heart failure  & 0.63 & 0.57 & 0.58 & 0.32 \\
Coronary atherosclerosis  & 0.62 & 0.58 & 0.52 & 0.31 \\
Shock  & 0.55 & 0.50 & 0.46 & 0.16 \\
Essential hypertension  & 0.61 & 0.61 & 0.61 & 0.43 \\
Hypertension  & 0.46 & 0.45 & 0.43 & 0.22 \\
\addlinespace[2pt]
\textit{Mean (8)}  & 0.54 & 0.50 & 0.46 & 0.25 \\
\midrule

\multicolumn{5}{l}{\textbf{Cerebrovascular}}\\
Acute cerebrovascular disease  & 0.61 & 0.60 & 0.12 & 0.07 \\
\addlinespace[2pt]
\textit{Mean (1)}  & 0.61 & 0.60 & 0.12 & 0.07 \\
\midrule

\multicolumn{5}{l}{\textbf{Renal}}\\
Renal failure  & 0.62 & 0.61 & 0.56 & 0.31 \\
Chronic kidney disease  & 0.48 & 0.45 & 0.44 & 0.23 \\
\addlinespace[2pt]
\textit{Mean (2)}  & 0.55 & 0.53 & 0.49 & 0.27 \\
\midrule

\multicolumn{5}{l}{\textbf{Respiratory}}\\
COPD  & 0.41 & 0.33 & 0.35 & 0.16 \\
Lower respiratory disease  & 0.24 & 0.17 & 0.20 & 0.11 \\
Upper respiratory disease  & 0.27 & 0.17 & 0.15 & 0.06 \\
Pulmonary collapse  & 0.34 & 0.22 & 0.22 & 0.10 \\
Pneumonia  & 0.52 & 0.45 & 0.42 & 0.18 \\
Respiratory failure  & 0.63 & 0.51 & 0.62 & 0.28 \\
\addlinespace[2pt]
\textit{Mean (6)}  & 0.40 & 0.31 & 0.33 & 0.15 \\
\midrule

\multicolumn{5}{l}{\textbf{Metabolic}}\\
$DM^*$ with complications  & 0.35 & 0.38 & 0.34 & 0.10 \\
$DM^*$ without complication  & 0.44 & 0.40 & 0.44 & 0.21 \\
Disorders of lipid metabolism  & 0.62 & 0.59 & 0.62 & 0.41 \\
Fluid and electrolyte disorders  & 0.67 & 0.66 & 0.65 & 0.44 \\
\addlinespace[2pt]
\textit{Mean (4)}  & 0.52 & 0.51 & 0.51 & 0.29 \\
\midrule

\multicolumn{5}{l}{\textbf{Gastrointestinal}}\\
Gastrointestinal hemorrhage  & 0.27 & 0.20 & 0.14 & 0.06 \\
Liver diseases  & 0.49 & 0.38 & 0.33 & 0.15 \\
\addlinespace[2pt]
\textit{Mean (2)}  & 0.38 & 0.29 & 0.24 & 0.10 \\
\midrule

\multicolumn{5}{l}{\textbf{Infectious}}\\
Sepsis  & 0.56 & 0.52 & 0.48 & 0.21 \\
\addlinespace[2pt]
\textit{Mean (1)}  & 0.56 & 0.52 & 0.48 & 0.21 \\
\midrule

\multicolumn{5}{l}{\textbf{Iatrogenic}}\\
Complications of medical care  & 0.46 & 0.42 & 0.43 & 0.22 \\
\addlinespace[2pt]
\textit{Mean (1)}  & 0.46 & 0.42 & 0.43 & 0.22 \\
\bottomrule
\multicolumn{5}{l}{\footnotesize *Diabetes mellitus}\\
\end{tabular}%
}
\end{table}

\subsection{Missing-Modality Performance Evaluation}
\label{sec:results_missing}

Table~\ref{tab:missing_modalities} evaluates our routing-based method under different modality combinations at inference time. Although the model is trained using all available modalities, we simulate realistic clinical settings at test time by masking the modality on the test data at inference time. Among unimodal models, structured data achieves the best performance for ICU mortality, indicating the strong predictive signal contained in vitals and laboratory measurements. In contrast, clinical notes yield the strongest performance for phenotype prediction, reflecting the richness of information captured in free-text documentation for more complex clinical characterization.

Among the bimodal combinations, $L+N$ performs best, indicating strong synergy between structured data and clinical notes. The trimodal setting achieves the highest overall performance, but performance under $L+N$ degrades less compared to $L+I$. This trend is consistent with the unimodal results, where both notes-only and structured-only models outperform the image-only model, highlighting the relative predictive strength of structured data and notes compared to imaging.

\begin{table}[!t]
\centering
\caption{Performance of our model compared across different modality combinations using AUROC and F1.  The model is trained on paired trimodal data and evaluated with modalities masked on test data. Best results are bolded.}
\label{tab:missing_modalities}
\begin{tabular}{lcc|cc}
\toprule
& \multicolumn{2}{c|}{Mortality} & \multicolumn{2}{c}{Phenotypes} \\
Modalities & AUROC  & F1  & AUROC & F1  \\
\midrule
$L$ only & $0.76_{\pm0.02}$ & $0.42_{\pm0.03}$ & $0.67_{\pm0.03}$ & $0.40_{\pm0.03}$  \\
$N$ only & $0.71_{\pm0.02}$ & $0.38_{\pm0.02}$ &$0.69_{\pm0.01}$ & $0.42_{\pm0.02}$ \\
$I$ only & $0.65_{\pm0.02}$ & $0.31_{\pm0.01}$ & $0.64_{\pm0.02}$ & $0.38_{\pm0.02}$ \\
\hline
$L{+}N$     & $0.74_{\pm0.02}$ & $0.39_{\pm0.02}$ & $0.71_{\pm0.02}$ & $0.44_{\pm0.02}$ \\ 
$L{+}I$    & $0.70_{\pm0.01}$ & $0.38_{\pm0.01}$ & $0.68_{\pm0.02}$  & $0.41_{\pm0.02}$ \\
\hline
$L{+}N{+}I$ & $\mathbf{0.80_{\pm0.02}}$ & $\mathbf{0.45_{\pm0.01}}$  & $\mathbf{0.78_{\pm0.01}}$ & $\mathbf{0.50_{\pm0.01}}$ \\
\bottomrule
\end{tabular}
\end{table}

\subsubsection{Phenotype Analysis}
At the phenotype level, performance degradation under missing modalities reveals clinically meaningful modality dependencies (Table \ref{tab:per_pheno_grouped_1col_auroc_f1_prev}). 

\begin{figure*}[t]
  \centering
  \vspace{-2mm}
  \includegraphics[width=\textwidth,height=0.46\textheight]{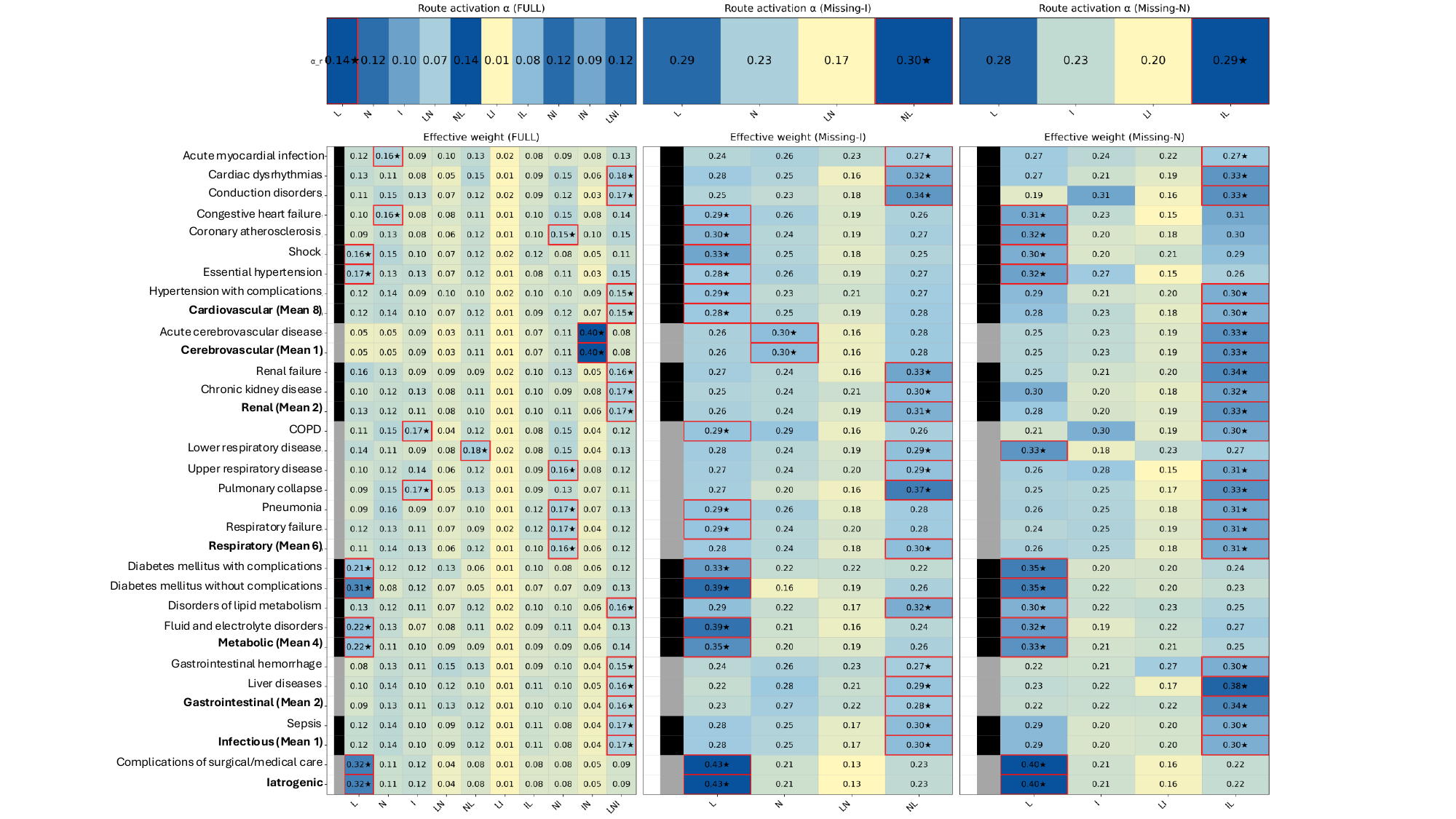}
  \vspace{-2mm}
  \caption{Routing activation and effective weight across all phenotypes.  The top heatmap shows the average primary activation ($\alpha_r$) across routing primaries.
  The bottom heatmaps show the effective routing weights for each phenotype.
  Three heatmaps at the top and bottom correspond to the FULL setting and missing-modality settings (Missing-I and Missing-N).}
  \label{fig:missing_heatmaps_N}
  \vspace{-3mm}
\end{figure*}

\begin{figure*}[t]
  \centering
  \includegraphics[width=\textwidth,height=0.12\textheight]{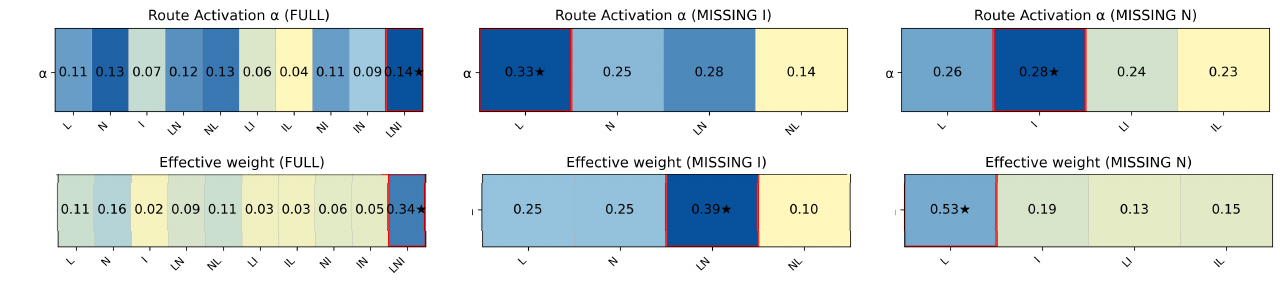}
  \caption{Routing activation and effective weight for mortality.  The top heatmap shows the average primary activation ($\alpha_r$) across routing primaries.
  The bottom heatmaps show the effective routing weights for mortality.
  Three heatmaps at the top and bottom correspond to the FULL setting and missing-modality settings (Missing-I and Missing-N).}
  \label{fig:app_mortality_primary_effective}
\end{figure*}

At the group level, cardiovascular, renal, infectious, and gastrointestinal phenotypes show a consistent pattern where performance drops more when clinical notes are missing compared to imaging, indicating that structured data and narrative information form the core predictive signals, while imaging plays a smaller role.

Within the cardiovascular and renal groups, some phenotypes, such as essential hypertension, cardiac dysrhythmias, and chronic kidney disease, remain largely unaffected by missing modalities, reflecting their primary reliance on structured data. In contrast, acute myocardial infarction, heart failure, shock, and coronary atherosclerosis exhibit moderate, symmetric declines when either notes or imaging are removed, suggesting that both modalities provide complementary information. 

A similar pattern is observed in iatrogenic phenotypes, where performance declines symmetrically when either notes or imaging is removed. This again indicates that predictions for these conditions rely on a balanced integration of multiple modalities.
 
Respiratory phenotypes exhibit overall high reliance on imaging, reflecting the importance of chest X-rays in this group compared to other phenotype categories. Despite this, the performance drop is similar under both missing images and missing notes scenarios, suggesting that while the model depends on imaging, the clinical notes containing radiology report details provide complementary evidence for images.

Metabolic phenotypes demonstrate the highest resilience to missing-modality scenarios, reflecting their strong reliance on structured data ($L$), such as laboratory values, which serve as the primary clinical indicators for conditions like diabetes and lipid disorders. 

Cerebrovascular phenotypes achieve high performance despite low prevalence, but exhibit poor robustness to missing clinical notes, with a sharp drop in F1, while performance remains largely unchanged when imaging is removed. This asymmetric sensitivity suggests reliance on a noisy or shortcut signal rather than stable multimodal evidence. Such behavior highlights the importance of missing-modality analysis as an audit mechanism, where inspecting routing weights can reveal over-reliance on specific modalities and guide closer examination of the model’s decision-making pathways.

\subsection{Global Interpretability Analysis}
Figures~\ref{fig:missing_heatmaps_N} and \ref{fig:app_mortality_primary_effective} summarize average route activations and effective weights across the test cohort for phenotype and mortality prediction, respectively. Route activations, averaged across the dataset, capture overall route expressiveness, while routing coefficients are learned per label and per sample, reflecting outcome-specific routing preferences. Effective weights are computed as the dot product of route activations and routing coefficients, representing the final contribution of each interaction to the prediction. Routing coefficients are reported in Appendix Figure \ref{fig:app:pheno_route} and \ref{fig:mort_route} for completeness, as their qualitative patterns largely overlap with the effective weights. The dot-product operation primarily rescales and redistributes weights without substantially altering the relative importance of dominant routes, making effective weights a more concise and interpretable summary for the main text.
\subsubsection{Route activations.} In both phenotypes and mortality tasks, unimodal routes ($L$ and $N$) consistently exhibit strong activations, indicating that each modality independently provides meaningful predictive signals. In addition, several cross-modal routes, particularly $NL$, $NI$, and $LNI$, are highly active, demonstrating that the model frequently integrates information across modalities to gather patient-level evidence.

Our bimodal directional interactions, implemented via cross-attention, further reveal systematic asymmetries. In each bimodal setting, one directional route is consistently more activated than its reverse. Specifically, under the $N$–$L$ combination, $NL$ dominates, under $N$–$I$, $NI$ shows higher activation, and under $L$–$I$, $IL$ is more active. This behavior is expected, as the model learns that rich note representations benefit from contextual enhancement provided by structured and imaging data. Notably, radiographic representations gain substantial value when conditioned on physiological context such as vitals and laboratory measurements, whereas structured features derive relatively limited additional information directly from visual patterns.
 
Under missing-modality scenarios, both tasks shift toward greater reliance on structured data. Phenotype prediction additionally assigns substantial weight to $NL$ and $IL$ routes, reflecting the increased complexity and heterogeneous evidence requirements of phenotype-level tasks.

\subsubsection{Effective weights.}

Routing patterns through effective weights reveal how the model integrates multimodal information for each prediction task, including 25 phenotypes (Figure \ref{fig:missing_heatmaps_N}) and mortality (Figure \ref{fig:app_mortality_primary_effective}). Overall, trimodal interactions receive the highest weights for both mortality and phenotype prediction, indicating a strong preference for jointly leveraging all available modalities. Among bimodal routes, $NL$ consistently stands out, reinforcing earlier findings from route-activation analysis that clinical notes play a central integrative role.

\begin{figure*}[t]
  \centering
  \includegraphics[width=\textwidth,height=0.28\textheight]{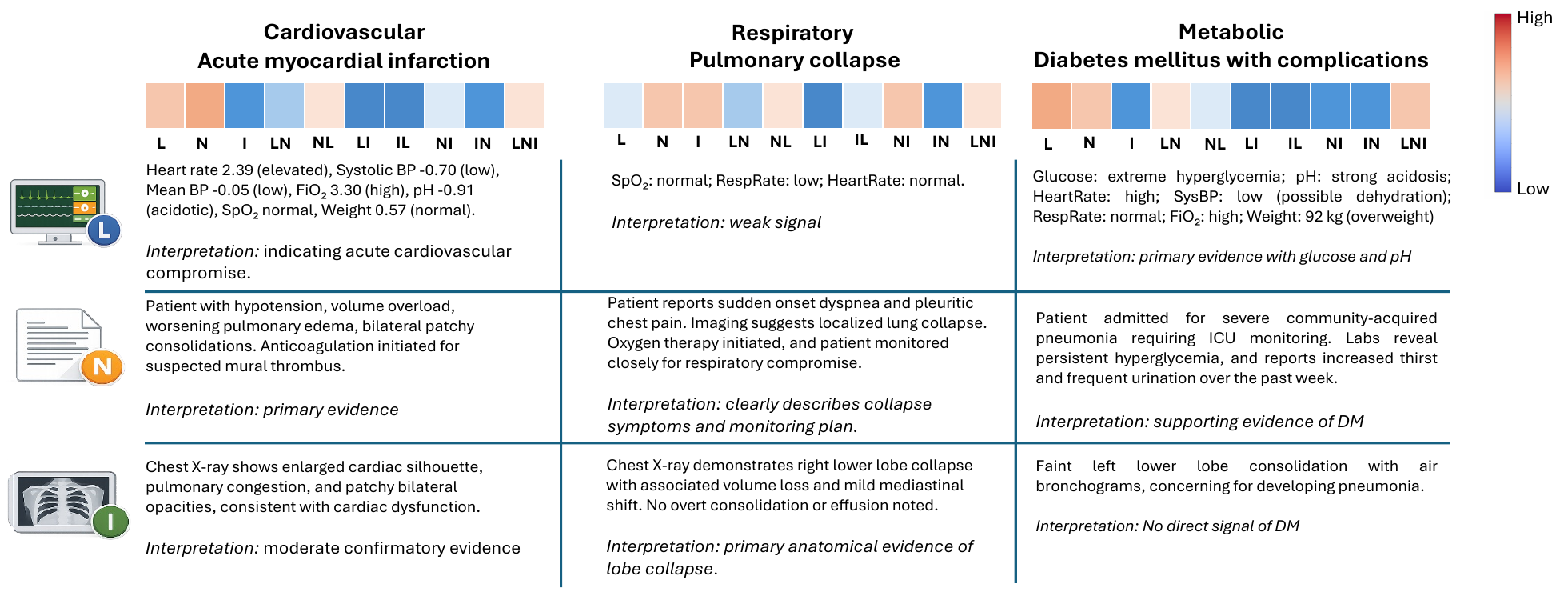}
  \caption{\textbf{The figure illustrates local interpretability for three patients, showing how routing weights are distributed across the three modalities for different phenotype prediction groups.}}
  \label{fig:local}
\end{figure*}

For phenotype prediction, when imaging ($I$) is missing, the most heavily weighted effective route is $NL$, reflecting reliance on structured data combined with narrative information. In contrast, under missing-notes ($N$) scenarios, weights are more evenly distributed across $L$ and $IL$, indicating a compensatory shift toward the remaining modalities. For mortality prediction, missing-modality scenarios lead to a stronger shift toward structured data ($L$) in both cases. However, when clinical notes ($N$) are available, weights are more evenly distributed across $L$, $N$, and $LN$, reinforcing the complementary importance of structured measurements and narrative information for mortality risk assessment.

\subsection{Group-level Interpretability}

As shown in Figure~\ref{fig:missing_heatmaps_N}, we group phenotypes into clinical categories and average effective routing weights within each group to characterize modality-reliance patterns. Combined with phenotype-level analyses, this provides a transparent view of how clinical signals are integrated at both phenotype and group levels, enabling systematic robustness analysis and auditing of multimodal integration.

Cardiovascular ($F_1=0.54$), renal ($F_1=0.55$), and infectious ($F_1=0.56$) phenotypes exhibit an integrated reasoning pattern, characterized by dominant reliance on the trimodal route $LNI$ (15–17\%) and secondary weighting on clinical notes $N$ (12–14\%). Consistent with this integration, these groups remain relatively robust under missing-modality scenarios, with performance drops of 7–11\% when imaging ($I$) is removed and 14–15\% when notes ($N$) are removed. When imaging is removed, routing shifts primarily toward structured data ($L$) and $NL$, with comparable weight assigned to notes, highlighting the central role of structured measurements and the added value of narrative context. Similarly, when notes are removed, weights shift toward $L$ and the strongest remaining multimodal route ($IL$), reflecting the importance of preserving multimodal interactions when possible. Directional bimodal preferences remain stable across missing-modality scenarios, while $LN$ and $LI$ remain consistently weak in their respective missing conditions, mirroring full-modality behavior.

Respiratory phenotypes also show balanced multimodal reliance, with increased contributions from imaging $I$ (9–17\%) while maintaining substantial weighting on clinical notes $N$ (11–16\%). Unlike other phenotype groups that primarily leverage $NL$ interactions, respiratory conditions show stronger reliance on $NI$ (16\%), reflecting the integration of chest X-ray evidence with narrative documentation, including radiology reports. Consistent with this routing behavior, performance drops most sharply when imaging is removed ($\Delta F_1=22.5\%$), highlighting the central role of radiographic evidence for respiratory phenotypes. When notes are removed, performance also declines substantially ($\Delta F_1=17.5\%$), indicating complementary dependence on narrative context. Under missing imaging, the model shifts weight toward structured data ($L$), clinical notes ($N$), and their interaction ($LN$). 
Conversely, under missing notes, routing redistributes across structured data ($L$), imaging ($I$), and their interaction ($IL$), reflecting phenotype-specific adaptive reweighting toward the strongest multimodal and unimodal pathways involving imaging.

On the other hand, gastrointestinal phenotypes show more skewed reliance on trimodal signals under full-modality conditions, with routing concentrated on $LNI$ and $LN$. Even in missing-modality scenarios, weights remain concentrated on the strongest available multimodal pathways, with limited shift toward structured data ($L$). Consistent with this behavior, performance drops sharply when either modality is removed ($\Delta F_1=23.7\%$ for missing imaging and $-36.8\%$ for missing notes), reflecting strong dependence on complementary multimodal information.

Metabolic ($F_1=0.52$) and iatrogenic ($F_1=0.46$) phenotypes both exhibit substantial reliance on structured data $L$ (22\% and 32\%, respectively), reflecting the dominant role of laboratory and vital measurements. For metabolic phenotypes, this reliance provides a stable core signal, with minimal degradation when notes or imaging are removed ($\Delta F_1=1.9\%$ for both). In contrast, iatrogenic phenotypes that show more skewed weighting towards $L$ (32\%), show larger performance drops ($\Delta F_1=8.7\%$ missing imaging and $-6.5\%$ missing notes), indicating that complementary multimodal signals are present but not effectively integrated. This suggests gaps in multimodal reasoning where the model fails to redistribute weights toward informative auxiliary modalities.

Cerebrovascular phenotypes assign substantial weight to a single route $IN$ (40\%) under full-modality input, yet exhibit asymmetric sensitivity to missing modalities. Performance drops minimally when imaging is removed ($\Delta F_1=1.6\%$) but collapses when notes are removed ($\Delta F_1=80.3\%$), with routing shifting to $IL$ and failing to sustain predictive accuracy. Conversely, removing imaging shifts the weight to $N$ while maintaining high performance, confirming that narrative notes are the primary predictive source. This suggests a potential shortcut effect, in which specific keywords in clinical notes ($N$) spuriously correlate with imaging signals ($I$), to support high performance in a full-modality scenario. 

These patterns in iatrogenic and cerebrovascular phenotypes illustrate how combining missing-modality analysis with routing weights provides effective audit signals. The observed performance drops and routing shifts reveal when multimodal evidence is underutilized or when the model relies on unstable or spurious correlations. This auditing framework therefore, exposes gaps in the model’s reasoning and highlights phenotypes in which additional training data, regularization, or architectural refinements are needed to better integrate complementary clinical evidence.

\subsection{Local interpretability}
Figure~\ref{fig:local} illustrates routing weights for three ICU patients across three phenotype prediction tasks: acute myocardial infarction, pulmonary collapse, and diabetes mellitus with complications. The figure highlights how our model dynamically distributes reliance across structured data ($L$), clinical notes ($N$), imaging ($I$), and their multimodal interactions.

For the patient with acute myocardial infarction, predictions are driven primarily by structured measurements (e.g., elevated heart rate, low systolic BP, acidosis) and relevant clinical notes describing hypotension, volume overload, and pulmonary edema. The trimodal route ($LNI$) receives moderate weight, reflecting confirmatory support from imaging, while other modality combinations contribute minimally.

In the pulmonary collapse patient, the model places high weight on clinical notes and imaging ($N$, $I$), with strong emphasis on their interaction ($NI$), capturing both the narrative description of collapse symptoms and radiographic evidence. Secondary contributions come from structured-to-note interactions ($NL$) and the full trimodal route ($LNI$), demonstrating the model’s balanced multimodal reasoning.

For diabetes mellitus with complications, the patient was admitted primarily for pneumonia, yet laboratory evidence (extreme hyperglycemia, acidosis) supports the DM phenotype. Here, the model relies heavily on structured data ($L$) and clinical notes ($N$), with lower weight to imaging. If the task shifts to pneumonia prediction, routing adapts to emphasize imaging and narrative evidence, illustrating task-specific interpretability.

Overall, these examples demonstrate that our framework enables transparent, patient-level reasoning and supports systematic auditing of evidence integration across phenotypes.

\section{Discussion}
We present a multimodal framework that explicitly models and exposes interactions across clinical modalities, enabling both strong predictive performance and interpretable reasoning pathways. Unlike conventional fusion approaches that implicitly mix signals, our routing mechanism disentangles unimodal, bimodal, and trimodal evidence contributions, providing a structured representation of how clinical information is integrated for each prediction.

Across phenotypes, the model demonstrates condition-specific reasoning strategies that align with clinical expectations, confirming that multimodal integration improves predictive accuracy through specialized reliance on informative signals. Importantly, routing weights offer a mechanistic explanation of model behavior, revealing whether predictions are supported by complementary evidence or dominated by a narrow signal that may be fragile or biased.

Missing modality experiments further validate that routing weights correspond to functional reliance. Performance degrades in clinically coherent ways when critical modalities are removed, indicating that the model is meaningfully integrating available information. Moreover, controlled redistribution of routing weights under modality ablation suggests graceful degradation rather than catastrophic failure, a key property for real-world clinical deployment where data incompleteness is common.

While the inherent difficulty of the task and class imbalances result in lower absolute F1 scores, our framework consistently outperforms all baselines in both AUROC and F1. A current limitation is that the number of directional bimodal routes scales quadratically with additional modalities ($O(M^2)$), which may increase computational overhead; future work will explore sparse or dynamic routing to extend this framework to higher-modality settings.

Overall, our framework advances multimodal clinical modeling by coupling strong predictive performance with explicit, multi-level interpretability. By exposing dominant evidence pathways and potential shortcut dependencies, it enables clinicians to assess reliability and identify failure modes. This positions multimodal routing as a practical mechanism for building transparent, trustworthy clinical decision-support systems.

\section{Conclusion}

We propose a multimodal routing framework that explicitly models unimodal and cross-modal evidence pathways. The model achieves competitive predictive performance while providing interpretable routing patterns that reveal disease-specific modality dependence. Combined with missing-modality experiments and routing explanations, this framework enables systematic auditing of multimodal clinical predictions and highlights when decisions are supported by robust multimodal evidence versus potentially unstable signals.

%\clearpage
\begin{acks}
NH, and ZW completed experiments. NH, EL and MG wrote the main manuscript text and prepared figures and tables. Computational resources were provided by the SMU O'Donnell Data Science and Research Computing Institute.
\end{acks}

\bibliographystyle{ACM-Reference-Format}

\bibliography{software}

@inproceedings{lee2023learning,
  title={Learning missing modal electronic health records with unified multi-modal data embedding and modality-aware attention},
  author={Lee, Kwanhyung and Lee, Soojeong and Hahn, Sangchul and Hyun, Heejung and Choi, Edward and Ahn, Byungeun and Lee, Joohyung},
  booktitle={Machine Learning for Healthcare Conference},
  pages={423--442},
  year={2023},
  organization={PMLR}
}

@article{johnson2023mimic,
  title={MIMIC-IV, a freely accessible electronic health record dataset},
  author={Johnson, Alistair EW and Bulgarelli, Lucas and Shen, Lu and Gayles, Alvin and Shammout, Ayad and Horng, Steven and Pollard, Tom J and Hao, Sicheng and Moody, Benjamin and Gow, Brian and others},
  journal={Scientific data},
  volume={10},
  number={1},
  pages={1},
  year={2023},
  publisher={Nature Publishing Group UK London}
}

@article{PhysioNet-mimic-iv-note-2.2,
  author = {Johnson, Alistair and Pollard, Tom and Horng, Steven and Celi, Leo Anthony and Mark, Roger},
  title = {{MIMIC-IV-Note: Deidentified free-text clinical notes}},
  journal = {{PhysioNet}},
  year = {2023},
  month = jan,
  note = {Version 2.2},
  doi = {10.13026/1n74-ne17},
  url = {https://doi.org/10.13026/1n74-ne17}
}

@article{PhysioNet-mimic-cxr-jpg-2.1.0,
  author = {Johnson, Alistair and Lungren, Matthew and Peng, Yifan and Lu, Zhiyong and Mark, Roger and Berkowitz, Seth and Horng, Steven},
  title = {{MIMIC-CXR-JPG - chest radiographs with structured labels}},
  journal = {{PhysioNet}},
  year = {2024},
  month = mar,
  note = {Version 2.1.0},
  doi = {10.13026/jsn5-t979},
  url = {https://doi.org/10.13026/jsn5-t979}
}

@article{cui2024multimodal,
  title={Multimodal fusion of ehr in structures and semantics: Integrating clinical records and notes with hypergraph and llm},
  author={Cui, Hejie and Fang, Xinyu and Xu, Ran and Kan, Xuan and Ho, Joyce C and Yang, Carl},
  journal={arXiv preprint arXiv:2403.08818},
  year={2024}
}

@article{wang2025missing,
  title={Missing-modality enabled multi-modal fusion architecture for medical data},
  author={Wang, Muyu and Fan, Shiyu and Li, Yichen and Xie, Zhongrang and Chen, Hui},
  journal={Journal of Biomedical Informatics},
  volume={164},
  pages={104796},
  year={2025},
  publisher={Elsevier}
}

@article{huang2020fusion,
  title={Fusion of medical imaging and electronic health records using deep learning: a systematic review and implementation guidelines},
  author={Huang, Shih-Cheng and Pareek, Anuj and Seyyedi, Saeed and Banerjee, Imon and Lungren, Matthew P},
  journal={NPJ digital medicine},
  volume={3},
  number={1},
  pages={136},
  year={2020},
  publisher={Nature Publishing Group UK London}
}

@article{harutyunyan2019multitask,
  title={Multitask learning and benchmarking with clinical time series data},
  author={Harutyunyan, Hrayr and Khachatrian, Hrant and Kale, David C and Ver Steeg, Greg and Galstyan, Aram},
  journal={Scientific data},
  volume={6},
  number={1},
  pages={96},
  year={2019},
  publisher={Nature Publishing Group UK London}
}

@inproceedings{lyu2023multimodal,
  title={A multimodal transformer: Fusing clinical notes with structured ehr data for interpretable in-hospital mortality prediction},
  author={Lyu, Weimin and Dong, Xinyu and Wong, Rachel and Zheng, Songzhu and Abell-Hart, Kayley and Wang, Fusheng and Chen, Chao},
  booktitle={AMIA Annual Symposium Proceedings},
  volume={2022},
  pages={719},
  year={2023}
}

@article{ruan2025towards,
  title={Towards accurate and reliable ICU outcome prediction: a multimodal learning framework based on belief function theory using structured EHRs and free-text notes},
  author={Ruan, Yucheng and Tan, Daniel J and Ng, See-Kiong and Huang, Ling and Feng, Mengling},
  journal={Journal of Healthcare Informatics Research},
  pages={1--42},
  year={2025},
  publisher={Springer}
}

@inproceedings{alsentzer2019publicly,
  title={Publicly available clinical BERT embeddings},
  author={Alsentzer, Emily and Murphy, John and Boag, William and Weng, Wei-Hung and Jindi, Di and Naumann, Tristan and McDermott, Matthew},
  booktitle={Proceedings of the 2nd clinical natural language processing workshop},
  pages={72--78},
  year={2019}
}

@article{li2020behrt,
  title={BEHRT: transformer for electronic health records},
  author={Li, Yikuan and Rao, Shishir and Solares, Jos{\'e} Roberto Ayala and Hassaine, Abdelaali and Ramakrishnan, Rema and Canoy, Dexter and Zhu, Yajie and Rahimi, Kazem and Salimi-Khorshidi, Gholamreza},
  journal={Scientific reports},
  volume={10},
  number={1},
  pages={7155},
  year={2020},
  publisher={Nature Publishing Group UK London}
}

@article{hooman2025equitable,
  title={Equitable Electronic Health Record Prediction with FAME: Fairness-Aware Multimodal Embedding},
  author={Hooman, Nikkie and Wu, Zhongjie and Larson, Eric C and Gupta, Mehak},
  journal={arXiv preprint arXiv:2506.13104},
  year={2025}
}

@inproceedings{hayat2022medfuse,
  title={MedFuse: Multi-modal fusion with clinical time-series data and chest X-ray images},
  author={Hayat, Nasir and Geras, Krzysztof J and Shamout, Farah E},
  booktitle={Machine Learning for Healthcare Conference},
  pages={479--503},
  year={2022},
  organization={PMLR}
}

@article{zhou2021radfusion,
  title={Radfusion: Benchmarking performance and fairness for multimodal pulmonary embolism detection from ct and ehr},
  author={Zhou, Yuyin and Huang, Shih-Cheng and Fries, Jason Alan and Youssef, Alaa and Amrhein, Timothy J and Chang, Marcello and Banerjee, Imon and Rubin, Daniel and Xing, Lei and Shah, Nigam and others},
  journal={arXiv preprint arXiv:2111.11665},
  year={2021}
}

@article{wang2025integrating,
  title={Integrating Multimodal EHR Data for Mortality Prediction in ICU Sepsis Patients},
  author={Wang, Yi and Li, Weihua},
  journal={Statistics in Medicine},
  volume={44},
  number={10-12},
  pages={e70060},
  year={2025},
  publisher={Wiley Online Library}
}

@inproceedings{wang2025robust,
  title={Robust Multimodal Learning for Ophthalmic Disease Grading via Disentangled Representation},
  author={Wang, Xinkun and Wang, Yifang and Liang, Senwei and Tang, Feilong and Liu, Chengzhi and Hu, Ming and Hu, Chao and He, Junjun and Ge, Zongyuan and Razzak, Imran},
  booktitle={International Conference on Medical Image Computing and Computer-Assisted Intervention},
  pages={447--456},
  year={2025},
  organization={Springer}
}

@article{sun2024outcome,
  title={Outcome prediction using multi-modal information: integrating large language model-extracted clinical information and image analysis},
  author={Sun, Di and Hadjiiski, Lubomir and Gormley, John and Chan, Heang-Ping and Caoili, Elaine and Cohan, Richard and Alva, Ajjai and Bruno, Grace and Mihalcea, Rada and Zhou, Chuan and others},
  journal={Cancers},
  volume={16},
  number={13},
  pages={2402},
  year={2024},
  publisher={MDPI}
}

@article{patil2025multimodal,
  title={Multimodal Decision Support System for Improved Diagnosis and Healthcare Decision Making},
  author={Patil, Aniket and Patil, Viraj and Sankpal, Sangram and Patankar, Tanuja S and Bhute, Harsha},
  journal={Journal of Biology and Health Science},
  year={2025}
}

@inproceedings{yao2024drfuse,
  title={Drfuse: Learning disentangled representation for clinical multi-modal fusion with missing modality and modal inconsistency},
  author={Yao, Wenfang and Yin, Kejing and Cheung, William K and Liu, Jia and Qin, Jing},
  booktitle={Proceedings of the AAAI conference on artificial intelligence},
  volume={38},
  number={15},
  pages={16416--16424},
  year={2024}
}

@inproceedings{wang2025moe,
  title={MoE-Health: A Mixture of Experts Framework for Robust Multimodal Healthcare Prediction},
  author={Wang, Xiaoyang and Yang, Christopher},
  booktitle={Proceedings of the 16th ACM International Conference on Bioinformatics, Computational Biology, and Health Informatics},
  pages={1--9},
  year={2025}
}

@article{al2025multimodal,
  title={Multimodal Representation Learning Based on Personalized Graph-Based Fusion for Mortality Prediction Using Electronic Medical Records},
  author={Al-Dailami, Abdulrahman and Kuang, Hulin and Wang, Jianxin},
  journal={Big Data Mining and Analytics},
  volume={8},
  number={4},
  pages={933--950},
  year={2025},
  publisher={TUP}
}

@inproceedings{wortwein2022beyond,
  title={Beyond additive fusion: Learning non-additive multimodal interactions},
  author={W{\"o}rtwein, Torsten and Sheeber, Lisa and Allen, Nicholas and Cohn, Jeffrey and Morency, Louis-Philippe},
  booktitle={Findings of the Association for Computational Linguistics: EMNLP 2022},
  pages={4681--4696},
  year={2022}
}

@inproceedings{tsai2020multimodal,
  title={Multimodal routing: Improving local and global interpretability of multimodal language analysis},
  author={Tsai, Yao-Hung Hubert and Ma, Martin and Yang, Muqiao and Salakhutdinov, Ruslan and Morency, Louis-Philippe},
  booktitle={Proceedings of the 2020 conference on empirical methods in natural language processing (emnlp)},
  pages={1823--1833},
  year={2020}
}

@inproceedings{he2016deep,
  title={Deep residual learning for image recognition},
  author={He, Kaiming and Zhang, Xiangyu and Ren, Shaoqing and Sun, Jian},
  booktitle={Proceedings of the IEEE conference on computer vision and pattern recognition},
  pages={770--778},
  year={2016}
}

@article{han2024fusemoe,
  title={Fusemoe: Mixture-of-experts transformers for fleximodal fusion},
  author={Han, Xing and Nguyen, Huy and Harris, Carl and Ho, Nhat and Saria, Suchi},
  journal={Advances in Neural Information Processing Systems},
  volume={37},
  pages={67850--67900},
  year={2024}
}

@article{yun2024flex,
  title={Flex-moe: Modeling arbitrary modality combination via the flexible mixture-of-experts},
  author={Yun, Sukwon and Choi, Inyoung and Peng, Jie and Wu, Yangfan and Bao, Jingxuan and Zhang, Qiyiwen and Xin, Jiayi and Long, Qi and Chen, Tianlong},
  journal={Advances in Neural Information Processing Systems},
  volume={37},
  pages={98782--98805},
  year={2024}
}

\FloatBarrier
\clearpage
\onecolumn

\appendix

\noindent
\begin{minipage}[t]{0.48\textwidth}
\vspace{0pt}

\section{Data Preprocessing}
\label{app:cohort_details}

\paragraph{Structured EHR data.}
We use 17 routinely collected ICU variables, including 5 categorical variables
(e.g., Glasgow Coma Scale components) and 12 continuous variables
(e.g., vital signs and laboratory measurements), following prior MIMIC benchmark
work \cite{harutyunyan2019multitask}. Variables are discretized at a fixed time
resolution and normalized using statistics computed from the training split.
After one-hot encoding categorical variables, each time step is represented as a
fixed-length feature vector.

\paragraph{Clinical notes.}
For each ICU stay, all available clinical notes within the specified observation
window are collected and concatenated.

\end{minipage}
\hfill
\begin{minipage}[t]{0.48\textwidth}
\vspace{0pt}

We tokenize the text using the BioClinicalBERT tokenizer and split long documents
into chunks of up to 512 tokens. Each chunk is encoded independently, and
chunk-level embeddings are averaged to obtain a single note embedding per ICU stay.

\paragraph{Chest X-ray images.}
Chest X-ray images are preprocessed following the MedFuse \cite{hayat2022medfuse}
pipeline. Images are resized and normalized before being passed to the image
encoder.

\section{Routing Coefficient Heatmaps}
\label{app:routing-heatmaps}

\noindent
We add Figure~\ref{fig:app:pheno_route} and Figure~\ref{fig:mort_route} to show
routing weights for phenotypes and mortality.

\end{minipage}

\vspace{0.8em}

\begin{center}

\includegraphics[
    width=\textwidth,
    height=0.56\textheight
]{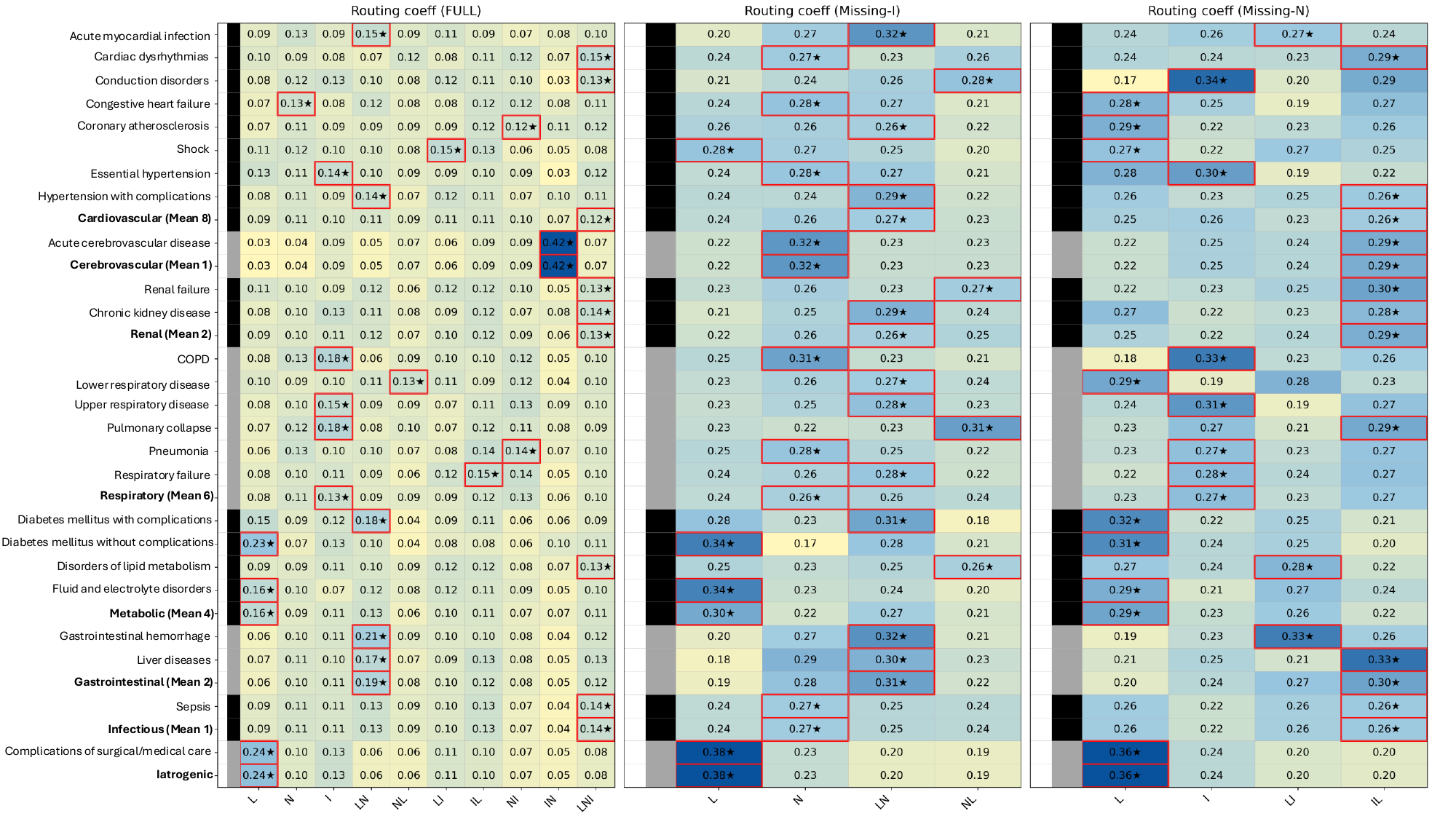}

\vspace{-0.6em}

\captionof{figure}{Routing coefficient heatmaps across all phenotypes. Three heatmaps correspond to the FULL setting and missing-modality settings (Missing-I and Missing-N).}
\label{fig:app:pheno_route}

\vspace{0.5em}

\includegraphics[
    width=\textwidth,
    height=0.10\textheight
]{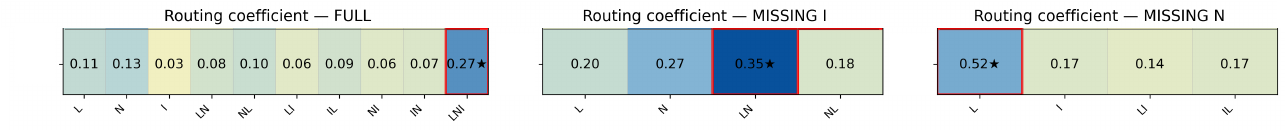}

\vspace{-0.6em}

\captionof{figure}{Routing coefficient heatmaps for mortality. Three heatmaps correspond to the FULL setting and missing-modality settings (Missing-I and Missing-N).}
\label{fig:mort_route}

\end{center}

\end{document}